\documentclass[letterpaper, 10 pt, conference]{ieeeconf} 
\IEEEoverridecommandlockouts 

\usepackage{times}
\usepackage{graphicx} 
\usepackage{subfigure} 
\usepackage{algorithm}
\usepackage{algorithmic}
\usepackage{hyperref}

\PassOptionsToPackage{options}{natbib}
\usepackage{mathtools}

\usepackage{todonotes}
\usepackage{times}
\usepackage{url}
\usepackage{amssymb}
\usepackage{amsmath}
\usepackage{chngcntr}
\usepackage{color}
\usepackage{lscape}
\usepackage[figuresleft]{rotating}
\usepackage{booktabs, topcapt}
\usepackage{nicefrac}       
\usepackage{microtype}      
\usepackage{paralist}

\newcommand{\half}{\nicefrac[]{1}{2}}
\newcommand{\threefour}{\nicefrac[]{3}{4}}
\newcommand{\twothree}{\nicefrac[]{2}{3}}

\newcommand{\be}{\begin{align}}
\newcommand{\ee}{\end{align}}
\newcommand{\benn}{\begin{align}}
\newcommand{\eenn}{\end{align}}
\newcommand{\la}{{\big \langle}}
\newcommand{\ra}{{\big \rangle}}

\title{\LARGE \bf
Disentangled Representations via Synergy Minimization
}

\author{Greg Ver Steeg$^{1}$, Rob Brekelmans$^{1}$, Hrayr Harutyunyan$^{2}$, and Aram Galstyan$^{1}$
\thanks{$^{1}$ University of Southern California, Information Sciences Institute
		{\tt\small gregv@isi.edu, brekelma@usc.edu, galstyan@isi.edu}}%
\thanks{$^{2}$ Yerevan State University, 
        {\tt\small hrayr@yerevann.com}}%
}

\begin{document} 

\bstctlcite{IEEEexample:BSTcontrol}
\bstctlcite{IEEEexample:ETAL}

\maketitle
\thispagestyle{empty}
\pagestyle{empty}

\begin{abstract}
Scientists often seek simplified representations of complex systems to facilitate prediction and understanding. If the factors comprising a representation allow us to make accurate predictions about our system, but obscuring any subset of the factors destroys our ability to make predictions, we say that the representation exhibits informational synergy. We argue that synergy is an undesirable feature in learned representations and that explicitly minimizing synergy can help disentangle the true factors of variation in underlying data. We explore different ways of quantifying synergy, deriving new closed-form expressions in some cases, and then show how to modify learning to produce representations that are minimally synergistic. We introduce a benchmark task to disentangle separate characters from images of words. We demonstrate that Minimally Synergistic (MinSyn) representations correctly disentangle characters while methods relying on statistical independence fail.
\end{abstract}

\section{Introduction} 

Instead of the colors ``blue'' and ``green'', imagine a language with the concepts of ``grue'', which describes something that is green during the daytime and blue at night, and ``bleen'', which is blue by day and green by night~\cite{goodman}.  
The Grue language looks unnecessarily complex to us, as we would have to describe a blue-eyed person as bleen-eyed by day and grue-eyed at night. 
However, one can imagine that the Grue-speaking people are familiar with two types of jellyfish: a poisonous, grue-colored species and a non-poisonous, bleen-colored species.
They would find our language inconvenient because of cumbersome warnings to avoid ``green during the day and blue at night''-jellyfish. Which set of concepts is simpler ultimately depends on the types of observations one is trying to characterize and this can be formalized with the concept of \textit{informational synergy}. 

Synergy is colloquially defined as a situation where the ``whole is more than the sum of its parts.'' To predict the current visual appearance of a green object from a description in the Grue language requires knowing whether the object is grue or bleen and whether it is currently night or day. Knowing either fact alone imparts no information. This mirrors the canonical example of synergy, the XOR gate on binary random variables with $X = Z_1 \oplus Z_2$, where either input, $Z_j$, alone is uninformative about $X$ while both together perfectly determine $X$.   

The focus of this work is to explore whether synergy can be useful in the context of unsupervised learning.  Just as different languages can exhibit more or less synergy while being equally expressive, latent variable models can exhibit more or less synergy while being equally predictive.  We thus introduce the principle of minimum synergy (MinSyn) for representation learning. 
We expect minimally synergistic representations to be more interpretable since each learned latent variable is encouraged to be individually informative about predicted observations.
Disentangling factors of variation is often cited as a goal for unsupervised learning~\cite{bengioreview}, but pinning down this hazy concept has proven difficult. Statistical independence of latent factors is often used as a proxy for disentangling~\cite{ica,nice}, but we demonstrate how independence-based approaches can fail to recover true structure while an approach based on synergy minimization succeeds. 

After reviewing various attempts to define a well-behaved information-theoretic measure of synergy \cite{griffith, mpisynergy, synergy_review, williamsbeer}, we identify a candidate measure which is well suited to our unsupervised representation learning problem and derive concrete formulations for the Gaussian and binary cases. 
We introduce an intuitive benchmark task where we train a model on  images of handwritten words and check whether factors are learned that correspond to individual characters. 
MinSyn learning correctly disentangles the characters while other methods fail.  We conclude with a discussion of open questions for using synergy to improve representation learning.

\section{Quantifying Synergy}\label{sec:quantify}

While there is no consensus on the correct way to measure synergistic information, we can understand how a group of random variables interact to predict a target variable using the Partial Information Decomposition framework~\cite{williamsbeer}.  For a given set of inputs $Z_{1:m} = \{Z_1, ..., Z_j,... Z_m\}$, we seek to understand how the mutual information $I(Z_{1:m}:X)$,with some target $X$, is distributed across predictors and subsets of predictors.  We seek to identify information, or reduction in uncertainty about the target $X$ \cite{cover}, as \textit{unique}, if it can be achieved only by conditioning on a single given predictor, \textit{redundant}, if it is available in more than one predictor, or \textit{synergistic}, if it appears only from considering interactions among two or more predictors.  Much of the literature to date on information decomposition has focused on finding measures that maintain a number of technical properties such as symmetry, monotonicity, and positivity, while also matching intuition across a variety of canonical examples (see \cite{griffith} for a detailed discussion). 

We now briefly introduce several candidate synergy measures and explore their properties before eventually deciding to use Correlational Importance, or CI synergy, to guide representation learning in Section \ref{sec:CI}. To compare measures and build intuition, Sec.~\ref{sec:gaussian_ex} describes a simple example of synergy for Gaussian variables. 



\subsection{Candidate Synergy Measures} \label{sec:synergy_measures} 
Among proposed measures of synergy, Whole Minus Sum synergy, arising in the neuroscience literature \cite{synergy}, directly formalizes the intuition of information in the whole minus information in the sum of the parts:
\begin{align}
\mbox{WMS}(Z_{1:m};X) \equiv I(Z_{1:m};X)  - \sum \limits_{j=1}^{m} I(Z_j;X)
\end{align}
This measure is maximized when each individual $Z_j$ contains no information about $X$, but taken together they contain full information about $X$. However, this quantity can be negative when there is redundant information that is counted multiple times in the sum over singleton predictors. Thus, by observing WMS synergy equal to zero, we cannot determine whether no synergy is present or if there is a mixture of synergy and redundancy~\cite{williamsbeer}.

To address these concerns, \cite{griffith} instead consider synergy as the ``whole minus union of the parts,''  where the union information is taken with respect to a variational distribution which solves the following optimization problem. 
\begin{align}
U(Z_{1:m};X) =~&  \underset{q(z_{1:m},x)}{\text{min}} I_q(Z_{1:m}:X) \label{eq:union} \\ 
&  \text{s.t. } q(z_j, x) = p(z_j,x), \forall z_j,x \nonumber
\end{align}
We use a subscript to denote mutual information with respect to the variational distribution, $q$. 
The marginal constraints on $q$ ensure that the information contained in individual predictors is unchanged. The minimization \textit{squeezes out} any information contained in higher order interactions among the $Z_j$. With this measure of union information, GK synergy is defined as the information contained in $Z$ that is not present in the union of information from individual $Z_j$'s, as follows. 
\begin{align}\label{eq:synergy}
S(Z_{1:m};X) \equiv I(Z_{1:m}:X) - U(Z_{1:m};X)
\end{align}

While this measure performs nicely for several low-dimensional exemplar problems and fits many theoretical desiderata \cite{griffith}, its applicability is limited by the intractable optimization in Eq.~\ref{eq:union} over all distributions consistent with the marginals, for which we have no closed form solution. 

To make GK synergy tractable for learning problems, we considered a simpler scenario where we restrict distributions to be Gaussian. In that case, we were able to derive the first closed form expression for GK synergy, which is shown in Appendix~\ref{sec:appendix_gk}. The synergy-minimizing distribution obtained ends up looking quite simple, $q_{GK}(x|z_{1:m}) \sim \mathcal{N} (\rho_{*}~ z_{*}, 1-\rho_{*}^2)$, where $Z_*$ denotes the predictor most correlated with $X$ and $\rho_{*}$ measures the Pearson correlation between $X$ and the predictor $Z_*$. In other words, if we have multiple predicted variables, $X_i$, synergy is minimized if each $X_i$ depends on only a single latent factor, $Z_{\mbox{parent}(i)}$. An immediate corollary provides an interesting theoretical connection between synergy minimization and representation learning: \emph{Gaussian latent tree models minimize (Gaussian) GK synergy} of latent factors, $Z_{1:m}$, with respect to observed variables, $X_i$.

Latent tree models are difficult to learn, and the constraint of having only one parent for each observed variable can be overly simplistic. 
Therefore, despite the theoretical appeal of GK synergy, for the purpose of optimal representation learning we are motivated to explore more practical measures. 

\subsection{Correlational Importance (CI) Synergy}\label{sec:CI}
We now consider Correlational Importance \cite{nirenberg, latham_synergy}, or CI synergy, as a promising and tractable candidate measure for use in representation learning.  
CI was originally presented to characterize the relationship between an experimental stimulus and observed neural responses resulting from the brain's noisy encoding procedure.  The measure reflects the extent to which the neural code stores information about a given stimulus in individual neurons versus in correlations among multiple neurons.

More concretely, \cite{nirenberg, latham_synergy} calculate CI synergy as the KL divergence between the true posterior of stimulus given responses and a distribution $p_{CI}(x|z_{1:m})$ which ignores correlations in the encoding process:
\begin{align}
C(Z_{1:m};X) = D_{KL}\Big(p(x|z_{1:m}) \big \| p_{CI}(x|z_{1:m})\Big) \label{eq:ci_synergy} \\
\quad p_{CI}(x|z_{1:m}) \equiv \frac{p(x) \prod_j p(z_j|x)}{p_{CI}(z_{1:m})}\, \label{eq:min_ci} \\
\quad p_{CI}(z_{1:m}) \equiv \sum_x p(x) \prod_j p(z_j|x) \nonumber
\end{align} 
CI synergy was shown by \cite{latham_synergy} and \cite{merhav1994} to provide an upper bound on the information loss to using the CI decoder rather than the true maximum likelihood decoder \cite{cover} in communicating over a noisy channel with input $X$ and output $Z_{1:m}$.  

A sufficient condition for CI to achieve a global minimum of zero is that the encoding distribution follows the conditional independence assumptions and $p(z_{1:m}|x) = \prod_j p(z_j|x)$.  As shown in \cite{amari_synergy}, this condition is not necessary and the minimum can also be attained for non-factorized distributions if, over the entire support of $X$:
\begin{align*}
p(z_{1:m}|x) = \frac{p(z_{1:m})}{p_{CI}(z_{1:m})} \prod \limits_j p(z_j|x)
\end{align*}
We will explore a learning framework under the condition of minimal CI synergy in Section \ref{sec:minsyn_learning}.



\begin{figure*}[htbp]
\vskip 0.05in
\begin{center}
\centerline{(a)\includegraphics[width=0.25\textwidth]{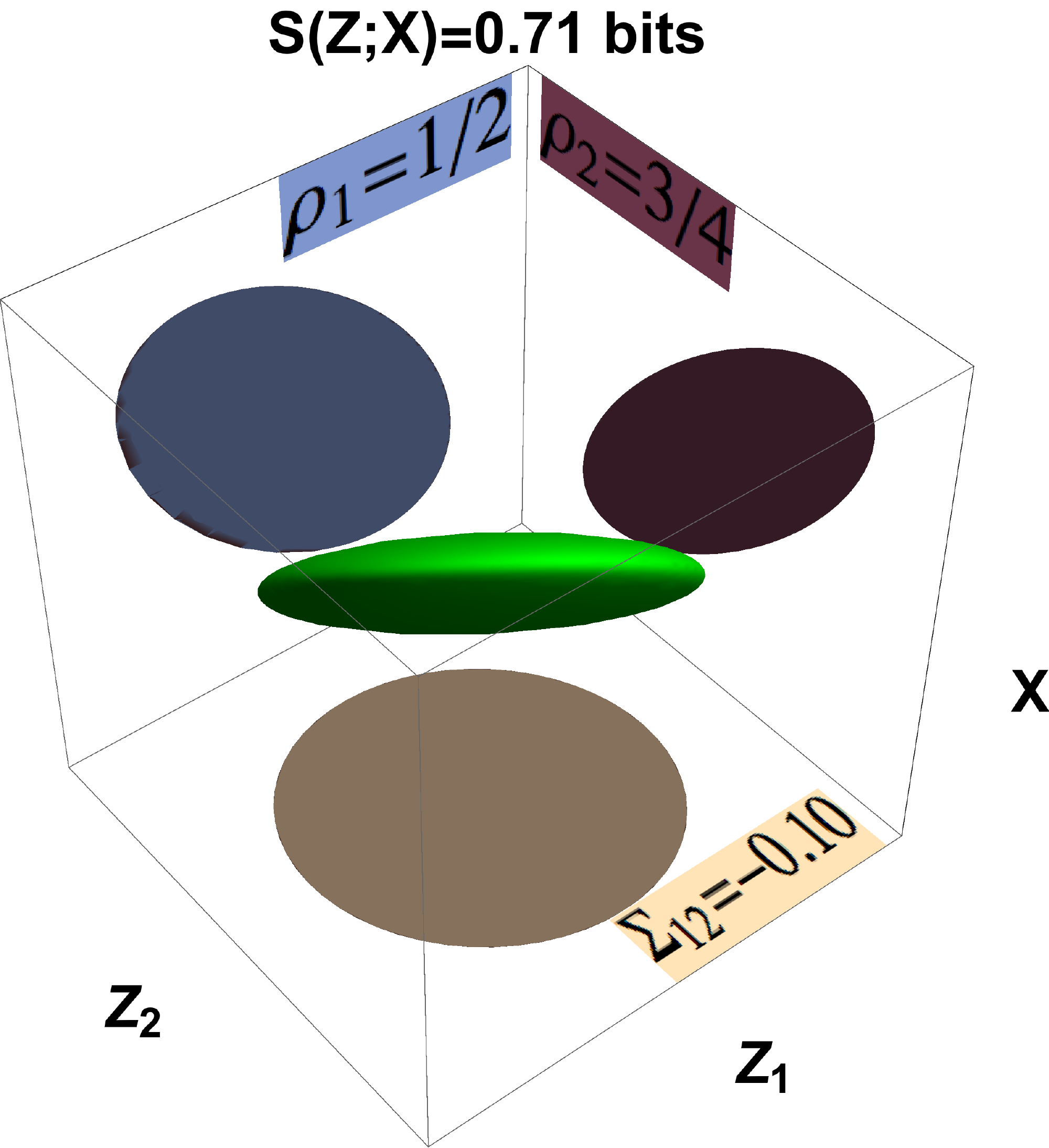}
(b)\includegraphics[width=0.25\textwidth]{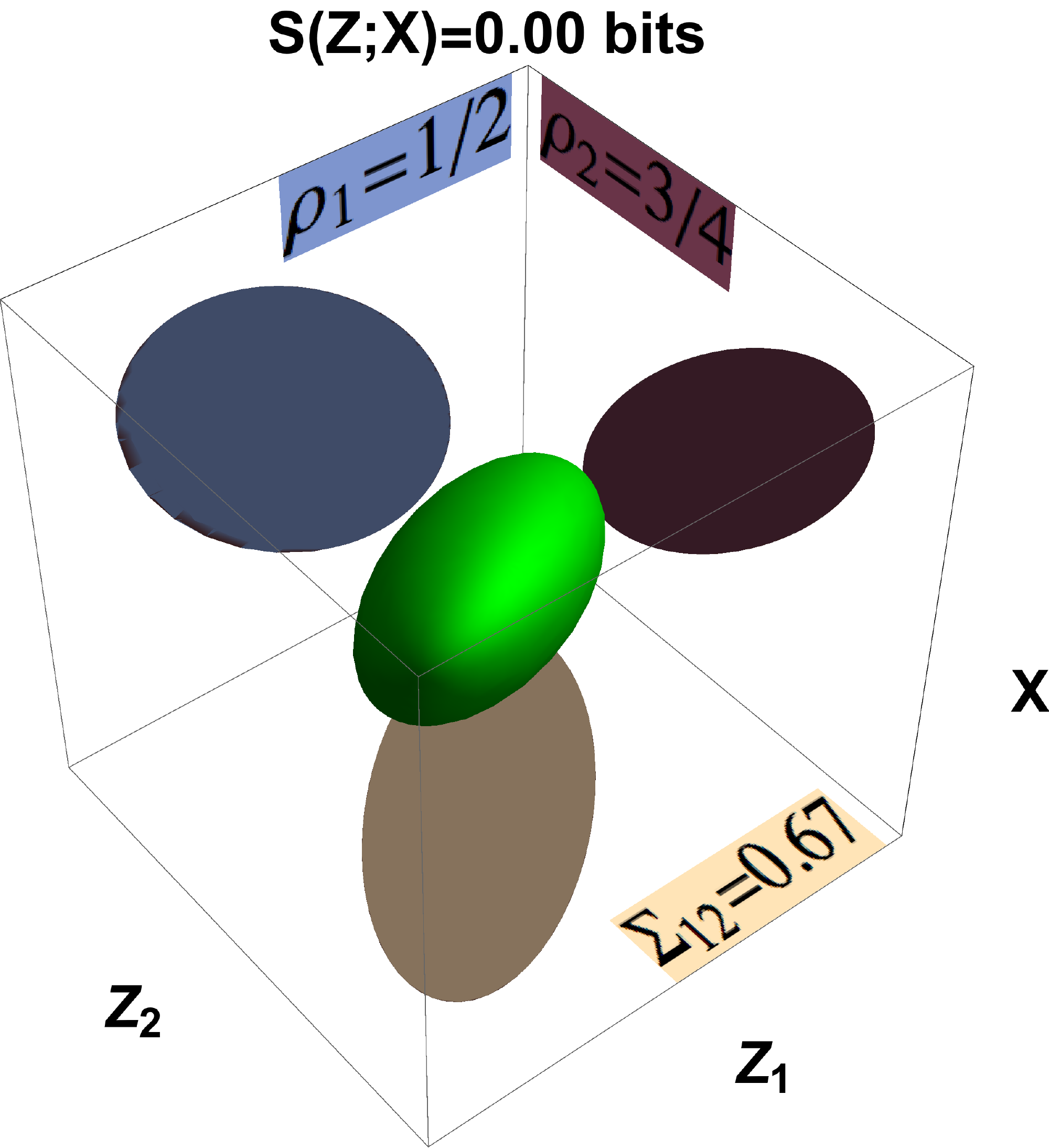} \qquad
(c)\raisebox{0.2\height}{\includegraphics[width=0.32\textwidth]{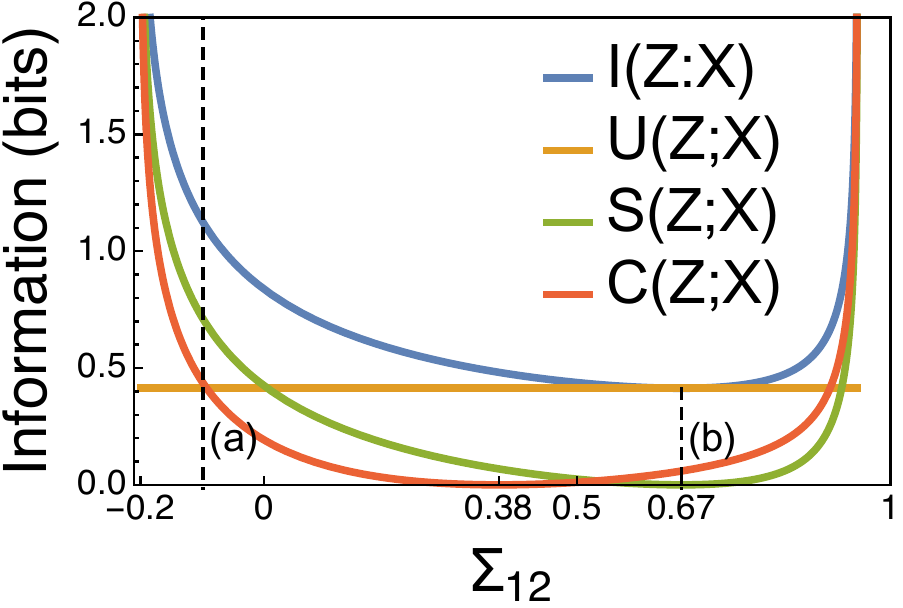}}}
\caption{For normal distributions over $Z_1,Z_2,X$, it is possible for $Z_1,Z_2$ to contain synergistic information about $X$. The ellipsoids represent level sets of normal distributions with covariance specified in Eq.~\ref{eq:sigma} along with the marginal projections for each pair of variables. Between (a) and (b), the value of $\Sigma_{1,2}$ changes while $\rho_1,\rho_2$ are unchanged. While (a) is synergistic, (b) exhibits zero GK synergy. (c) We show how mutual information, union information, GK synergy, and CI synergy change over the entire range of $\Sigma_{1,2}$. }
\label{fig:example}
\end{center}
\vskip -0.2in
\end{figure*}

\subsection{An Example of Gaussian Synergy} \label{sec:gaussian_ex}
Given the simple linear structure of a multivariate Gaussian, it may be surprising that these models can exhibit informational synergy.  For a three dimensional joint distribution over $Z_1, Z_2,$ and $X$, what types of correlation structures will lead to $Z_1$ and $Z_2$ being synergistic with respect to $X$?

Mutual information is invariant under scaling and translation of the marginals, so w.l.o.g. we take each variable to have a mean of zero and a standard deviation of one. We use brackets to indicate expectation values, defining $\rho_j = \la Z_j X\ra$ and $\Sigma_{1,2} = \la Z_1 Z_2\ra$. The marginal distributions between $X$ and each $Z_j$ are thus fully specified as 
$$p(z_j, x) \sim \mathcal N\left(\vec 0, \begin{pmatrix}
  1 &  \rho_j \\
  \rho_j & 1 
 \end{pmatrix} \right)$$

Further, the set of joint distributions compatible with these marginal relationships can be written as $q(z_1, z_2, x) \sim \mathcal N(\vec 0, \Sigma)$, where the covariance matrix has the form
\begin{align}\label{eq:sigma}
\Sigma = \begin{pmatrix}
  1 &  \Sigma_{1,2} & \rho_1\\
  \Sigma_{1,2} & 1 & \rho_2\\
  \rho_1& \rho_2& 1
 \end{pmatrix} 
 \end{align}
 
For this to specify a valid covariance matrix, it must be positive semidefinite which implies the constraints $\rho_j^2 \leq 1, \Sigma_{1,2}^2 \leq 1$, and the less trivial constraint that $\Sigma_{1,2}$ is between $ \rho_1 \rho_2 \pm \sqrt{(1-\rho_1^2)(1-\rho_2^2)}$. 

If we choose the $X$-marginals to be $\rho_1=\half$ and $\rho_2=\threefour$, the space of joint distributions is then parametrized only by the correlation between $Z_1$ and $Z_2$, $\Sigma_{1,2}$. We visualize the joint induced by two different choices of $\Sigma_{1,2}$ in Figure ~\ref{fig:example}: a synergistic case with $\Sigma_{1,2} = -.10$ in  \ref{fig:example}(a), and a non-synergistic case with $\Sigma_{1,2} = .67$ in \ref{fig:example}(b).  In both cases, the ellipsoids represent the level sets of the distribution, with the green representing the joint, and the vertical axis representing the target $X$.

For the synergistic case in Figure ~\ref{fig:example}(a), observe that, specifying the value of one of the predictors, $Z_1$ for example, leaves a plane of possible distributions for the other two variables.  The joint distribution is a tilted plate, so the value of $X$, or the coordinate of intersection with this plane on the vertical axis, is not specified by $Z_1$ alone.  On the other hand, knowledge of \textit{both} predictors gives a tight estimate of $X$ since the plate is thin,  and we thus expect any candidate measure to give non-zero synergy. 

In the non-synergistic case of Figure ~\ref{fig:example}(b), the ellipsoid for the joint distribution is fatter and more symmetric, meaning that our uncertainty about $X$ is similar whether conditioning on a single predictor or both together.  Indeed, there is zero GK synergy in this case.  We also plot how GK synergy and CI synergy 
vary with $\Sigma_{1,2}$ in Figure ~\ref{fig:example}(c). While areas of low and high synergy are qualitatively similar, the synergy-minimizing distribution, which is the main interest of this paper, differs for the two synergy measures.

\subsection{Independence versus Synergy}
Statistical independence can be quantified with total correlation  \cite{watanabe}, a multivariate generalization of mutual information.  Dependence among latent variables $Z_{1:m} = \{Z_1, Z_2, ..., Z_m \}$ is characterized:
\begin{align}
TC(Z_{1:m}) &= \sum \limits_{j=1}^{m} H(Z_j) - H(Z_{1:m}) 
\end{align}
This quantity is zero if and only if all $Z_j$ are independent, which can also be seen by a simple rearrangement of the entropy terms to obtain a KL divergence $D_{KL}(p(z_{1:m})|| \prod_j p(z_j))$.   Minimizing dependence between latent variables is a commonly proposed method for achieving disentangled representations \cite{ica, nice, bell95, informationdropout}.

Representations comprised of statistically independent latent factors differ from ones with minimal synergy. Consider the classic example of the XOR gate, where we take $Z_1, Z_2$ to be iid random variables taking binary values with probability half, and $X = Z_1 \oplus Z_2$. By definition $Z_1, Z_2$ are independent, meaning $TC(Z_{1:2}) = 0$, but all candidate measures for the synergy of $Z_{1:2}$ with respect to the prediction target, $X$, take on a maximal value of 1 bit~\cite{griffith}. Differences in learned representations are illustrated in Sec.~\ref{sec:results}. 




\section{MinSyn Learning} \label{sec:minsyn_learning}

For representation learning, we must shift from considering a one-dimensional target $X$ to a high-dimensional set of input features $X_{1:n} = \{X_1,..., X_i,..., X_n\}$.  We consider a standard auto-encoder architecture~\cite{bengio_autoencoders}, with encoding functions $Z_j = f_j(X_{1:n})$ transforming the data into some latent factor space, and decoding functions $\bar X_i = g_i(Z_{1:m})$ that approximately recover observations from latent representations. The functions can be parametrized, for instance, as neural networks.  Common training objectives for auto-encoders minimize a reconstruction loss between the observed variables, $X_i$, and the reconstructed ones, $\bar X_i$~\cite{goodfellow2016deep}.  While the optimal decoder with respect to the posterior probability of $X_i$ given $Z_{1:m}$ would be $g_i(z) = \mathbb E_p(X_i | Z_{1:m} = z)$, this is typically intractable. Thus, we replace the expectation over $p$ with a variational distribution $q$, which we will take to minimize the synergy among the latent factors $Z_{1:m}$ with respect to each predicted variable, $X_i$. 

We frame the synergy minimization problem over variational distributions $q(x_i,z_{1:m})$ which approximate the true distribution, $p$, as follows.
\begin{align}
\arg \min_{q(x_i|z_{1:m})} \mbox{Synergy}_q(Z_{1:m};X_i) \label{eq:syn_opt}\\  \text{s.t.}~ \forall j \, \quad q(x_i, z_j) = p(x_i,z_j) \nonumber
\end{align}
The marginal constraint ensures that the information that each individual latent factor contains about a predicted variable is unchanged, while the information contained jointly among latent factors is tuned to minimize a synergy measure.

For mathematical tractability, we choose to minimize CI synergy. 
Solving the optimization in Eq.~\ref{eq:syn_opt} using the CI synergy measure in Eq.~\ref{eq:ci_synergy}, we see that the global minimizer of CI $= 0$ is achieved for the distribution $q(x_i|z_{1:m}) = p_{CI}(x_i|z_{1:m})$ which satisfies the constraints of our optimization problem and depends only on marginal relationships between $X_i, Z_j$.
 
Given this framework for MinSyn learning, we just need to obtain an explicit form for the decoder which minimizes CI synergy.  We provide expressions for MinSyn decoders for the Gaussian and binary cases below.

\subsection{CI Synergy for Gaussian Decoders}\label{sec:gauss}
 
Assume the relationships between each latent variable $Z_j$ and $X_i$ are Gaussian. For simplicity and w.l.o.g. we give expressions below for variables that have been standardized to have zero mean and unit variance.
The form of the CI minimizing distribution, $p_{CI}(x_i|z_{1:m})$, is given in Eq.~\ref{eq:min_ci}. 
The distribution depends on the correlations $\rho_{ij} \equiv \la X_i Z_j \ra$. 
\begin{align}\label{eq:ci_gauss}
p_{CI}(x_i| z_{1:m}) &= \mathcal N\left(\frac{1}{1+R_i} \sum_j \frac{\rho_{ij}}{1-\rho_{ij}^2} z_j, \frac{1}{1+R_i}\right)
\end{align}
where $R_i = \sum_j \frac{\rho_{ij}^2}{1-\rho_{ij}^2}$.  Now the CI synergy in Eq.~\ref{eq:ci_synergy} is just defined as the KL divergence between two Gaussian distributions, the actual distribution $p(x_i|z_{1:m})$ and the distribution assuming conditional independence above.


\subsection{CI Synergy for Binary Decoders}

In Sec.~\ref{sec:results} we will also show results where outputs are considered as binary variables, so we give an expression for a MinSyn decoder where all variables are binary. As above, this takes the form of  $p_{CI}(x_i=1|z_{1:m})$, which implies the following form for the decoder distribution.
\begin{align}
p_{CI}(x_i=1|z_{1:m}) = \mbox{sig}(\sum_j  w_{i,j}z_j + b_i) \label{eq:decoder}
\end{align}
where the sigmoid function, often used in neural networks, is defined $\mbox{sig}(v) = 1/(1+e^{-v})$ and $w, b$ denote weights and biases respectively~\cite{goodfellow2016deep}.  Instead of explicitly training $w, b$ as in a typical neural network, these parameters are now set according to the joint statistics of the observations and encoded states as follows. 
\begin{align*}
b_i &= \log \frac{p(X_i = 1)}{p(X_i=0)} + \sum_j \log \frac{p(Z_j=0|X_i=1)}{p(Z_j=0|X_i=0)} \\
w_{i,j} &= \log \frac{p(Z_j=1|X_i=1) p(Z_j=0|X_i=0)}{p(Z_j=0|X_i=1) p(Z_j=1|X_i=0)}
\end{align*}

In both the binary and Gaussian cases, our MinSyn decoder depends on the statistics of the pairwise marginals. It would be time consuming to compute these for the entire dataset after each gradient update. Therefore, we follow a similar strategy to that used in Batch Normalization~\cite{batchnorm}. During training, we calculate statistics only on the batch of samples currently being used for a gradient update. At the same time, we update a moving average of these statistics. During test time, we used the learned average.

\section{Results}\label{sec:results}

\subsection{EMNIST Handwritten Words Dataset}
We begin by introducing a new dataset and benchmark task for disentangling natural factors of variation. 
A common task in representation learning is to construct low-dimensional latent factor representations of handwritten characters (like the famous MNIST digits) that can be used to reconstruct the original images. 
We extend this problem to consider combinations of handwritten letters into words. What is the most natural (disentangled) way to represent this type of data? 
One way to represent this space is to include a latent factor for each possible word. However, the number of possible words is large so representing the data in this way could be inefficient. Moreover, if new words are introduced, we will need to add new factors and retrain the model. A more succinct representation would learn about the existence of characters and represent words as combinations of characters. The number of factors required is then just the number required to represent a character times the number of characters in the word. Therefore, we hypothesize that character disentangled representations will be better able to reconstruct new words that were not seen in training. 

We consider representations that recover a character-level representation from raw image data to have correctly disentangled the natural factors of variation. 
It is common in computer vision to build in assumptions of spatial locality using convolutional layers and pooling. Since characters are trivially distinguished through spatial localization, we can only fairly use this benchmark to test disentangling on domain-agnostic methods. Knowledge of the spatial layout of pixels cannot be used. 

ICA and latent factor generative models typically assume independent latent factors. For this example, it is clear that the character-level factors are not independent since not all combinations of characters result in valid English words. For instance if you see a word that starts with ``t'', the second letter is much more likely to be ``h'' than any other character due to the ubiquity of the word ``the'' in the English language. Therefore, we hypothesize that disentangling based on independence will perform poorly for this benchmark task. 

\begin{figure}[htbp]
\vskip 0.07in
\begin{center}
\includegraphics[width = 0.8 \columnwidth]{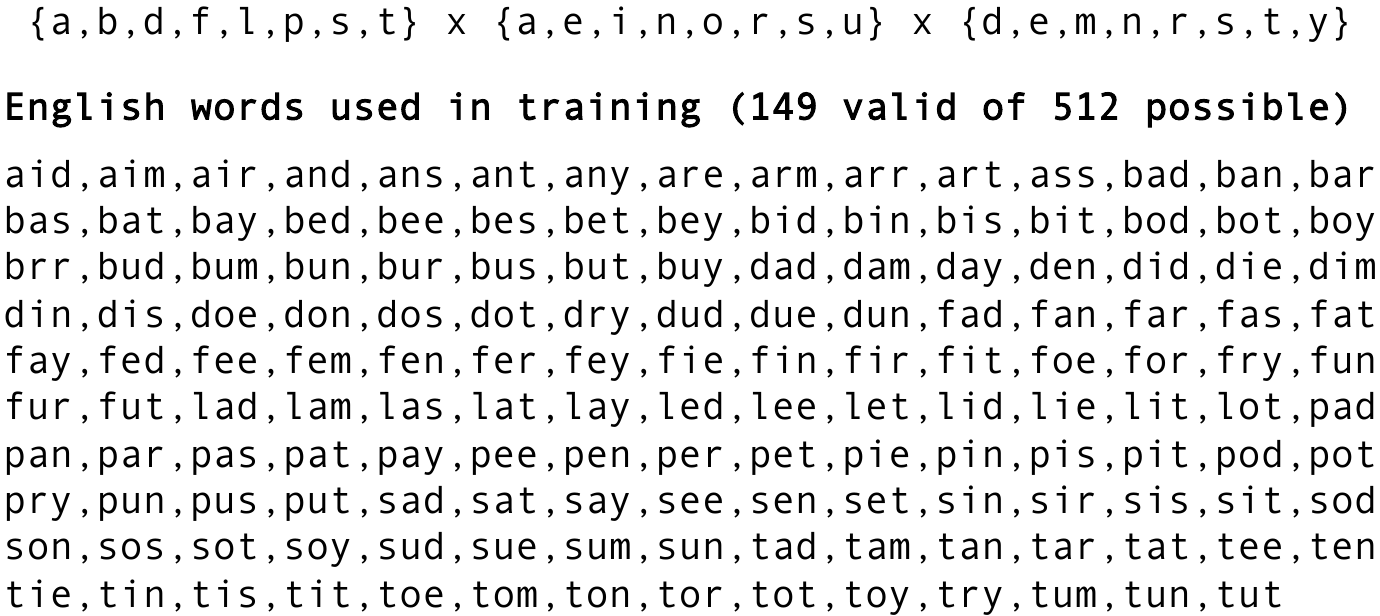}
\caption{Characters and words used to construct the training set.} \label{fig:word_list}
\end{center}
\vskip -0.2in
\end{figure} 
For the current experiments, we construct a simple dataset of handwritten word images composed of letters from the EMNIST dataset \cite{cohen2017emnist} (see Figure \ref{fig:word_recon} for several examples).  We first observe the 300 most commonly appearing English words, and consider the 8 letters that appear most frequently in the first, second, and third positions of these words.  Any of the 512 combinations of these letters which form valid English words are included as a training sample, leading to a dataset of 149 word images (Fig.~\ref{fig:word_list}).  We construct a test set using all remaining combinations of these letters, noting that reconstruction performance on this set should be hampered if our representations memorize words instead of learning individual letters.  In this experiment, we use identical image samples each time a letter appears in a word to remove variation in capitalization and writing style, an assumption that would match a typewritten text scenario. This variation could be included for added challenge in future experiments. 

Code constructing this dataset can be found at \cite{github}, and can be used as a template for exploring novel ways of sampling words and letters.  Just as in the Grue example, the ideal disentangled representation may vary depending on details of the relevant task.

\subsection{Disentangled Representations from Handwritten Words}\label{sec:disentangling}
We now apply several latent factor learning methods to the dataset of handwritten words generated as described in the previous section. For each method, we use 9 latent factors under the intuition that if a disentangled representation uses 3 latent factors to describe each character position, then these three factors will be capable of distinguishing $2^3=8$ different characters in that position. 

We considered several approaches including two methods based on independence and two methods based on minimizing synergy. The independence-based methods we used were FastICA \cite{ica} and InfoMax \cite{bell95}, implemented using \cite{scikit} and \cite{Gramfort2014446}, respectively. The MinSyn implementations are based on the derivation of MinSyn decoders in the Gaussian and binary cases in Sec.~\ref{sec:minsyn_learning}. We also compare to an auto-encoder with no synergy constraint. Finally, a common trick to improve robustness of standard auto-encoders is to add Gaussian noise to the inputs in training, a so-called denoising auto-encoder. Implementations are described in detail in Sec.~\ref{sec:implementation}.

\subsubsection{Disentangled latent factors}
\begin{figure}[tbp]
\vskip 0.07in
\begin{center}
\includegraphics[width = .45\textwidth]{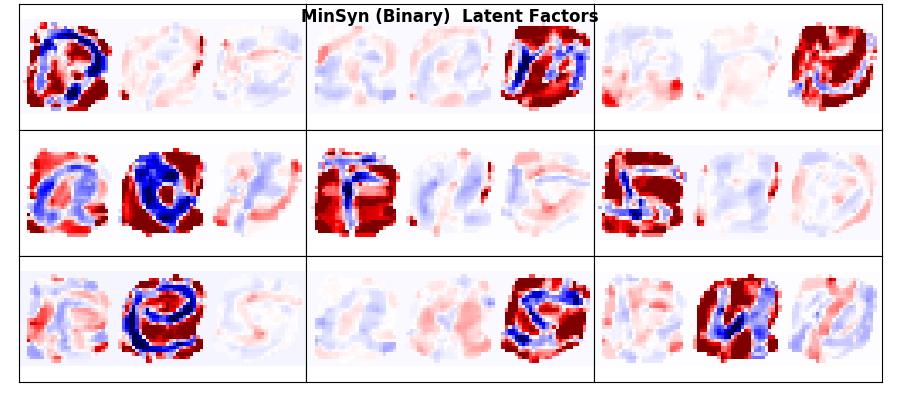}
\includegraphics[width = .45\textwidth]{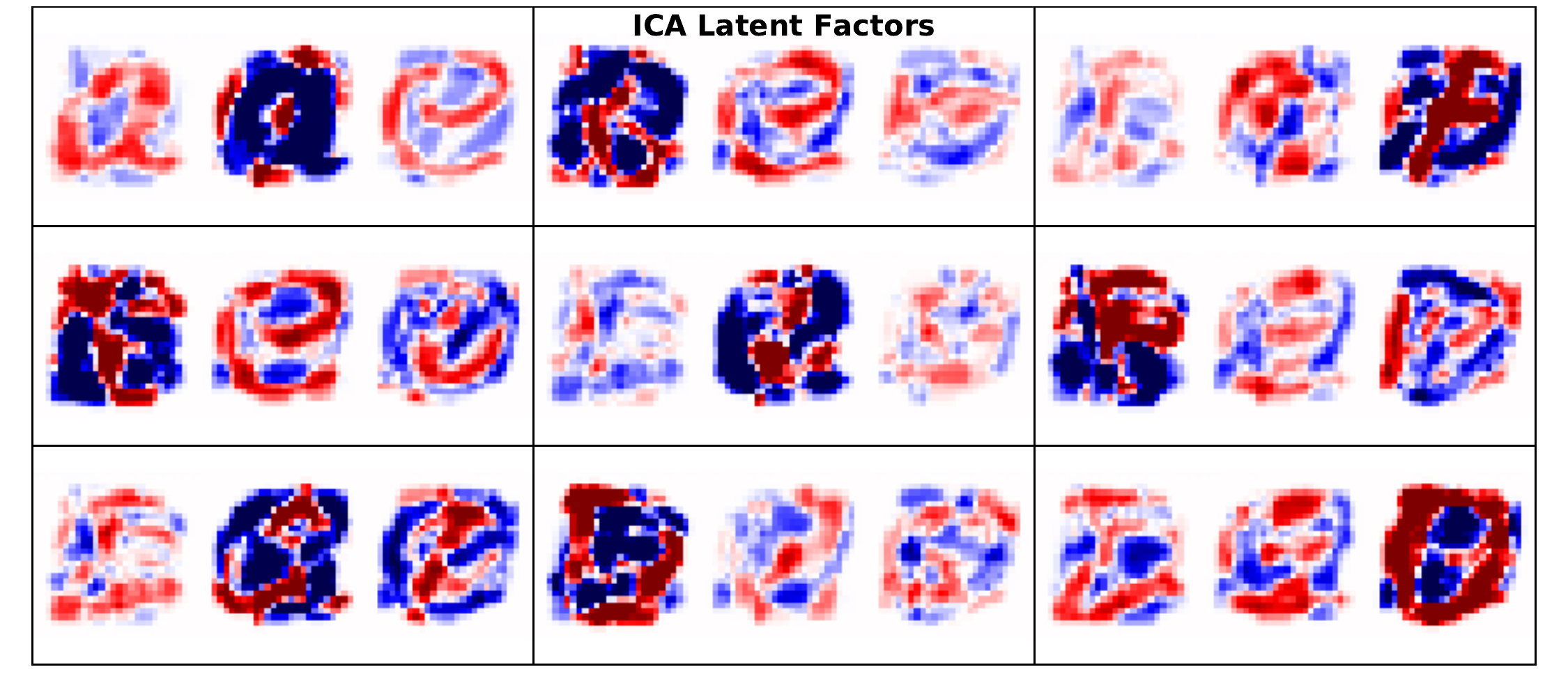}
\includegraphics[width = .45\textwidth]{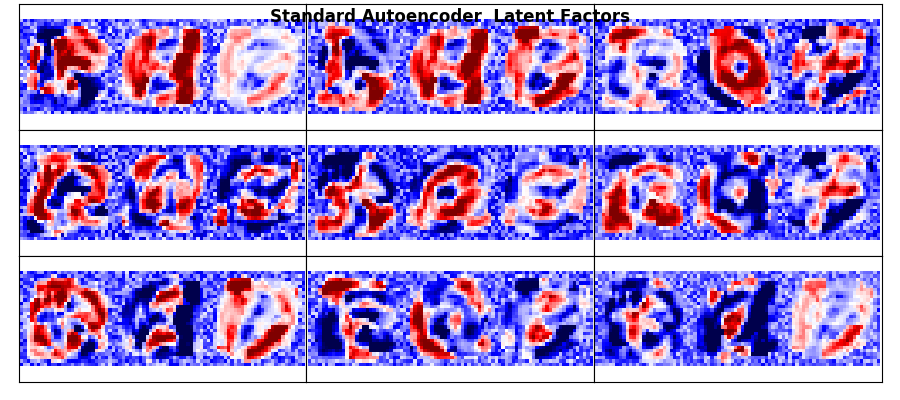}
\caption{Decoder weights for each of 9 latent factors learned by i) MinSyn Binary, ii) FastICA, and iii) an auto-encoder without synergy constraints} \label{fig:disent}
\end{center}
\vskip -0.2in
\end{figure}

In Figure \ref{fig:disent}, we visualize decoder maps learned by different methods. Each method has a decoder map for each of 9 latent factors, presented in no particular order. Our expectation for disentangled representations is that each latent factor should correspond to only a single character, and therefore the associated decoder map will have nonzero weight in only one character position. 


To quantify the overlap in decoder feature maps, we define the average character concentration (ACC) score and report it in Table~\ref{tab:result}. For a given factor $j$ and pixel $i$, we have decoder weights, $w_{i,j}$. We denote the set of pixels that belong to character $k$ as $S_k$. We define the concentration of weights in character $k$ as $C_{j,k} = \sum_{i \in S_k} w_{i,j}^2/ \sum_{i} w_{i,j}^2$. Finally, ACC is defined as the average entropy across $m$ latent factors, $ACC = 1/m \sum_j \sum_k -C_{j,k} \log C_{j,k}$. ACC would be zero if each latent factor were concentrated on an individual character position.

Figure \ref{fig:disent} shows that standard auto-encoders produce very entangled representations and looking at the ACC scores in Table~\ref{tab:result} shows that denoising does not help.  
As hypothesized, the dependence between characters induced by sampling English words causes entanglement of the ICA sources, with learned features spanning multiple positions. InfoMax fares even worse than ICA both visually (not shown) and according to ACC score. 
While statistical independence at first glance seems to be a natural way of quantifying disentanglement, this benchmark exemplifies how this intuition can fail. 
In contrast, MinSyn learns qualitatively different maps that both visually disentangle the characters and have lower ACC scores than any other methods.


\begin{figure}[tbp]
\vskip 0.07in
\begin{center}
{\small Training data (English words) and reconstructions} \\
\includegraphics[trim=0 50 0 0,clip,width = .45 \textwidth]{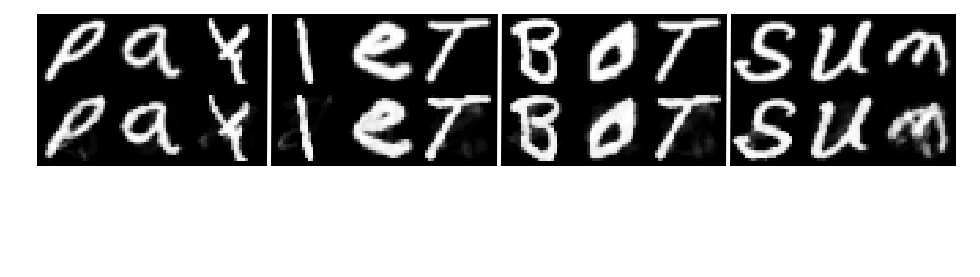} \\
{\small Test data (not words) and reconstructions} \\
\includegraphics[trim=0 50 0 0,clip,width = .45 \textwidth]{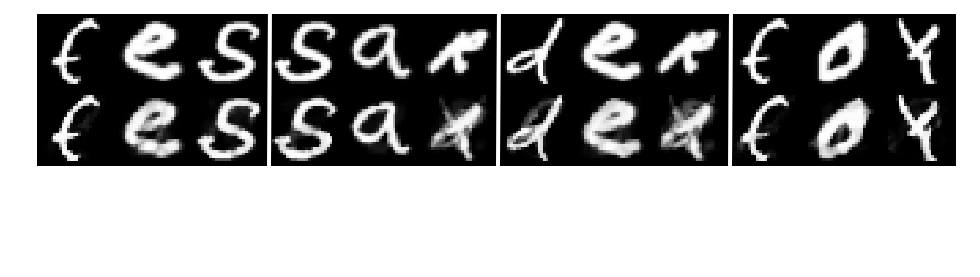} 
\caption{Train/test data shown in top row, with MinSyn reconstruction below.} \label{fig:word_recon}
\end{center}
\vskip -0.2in
\end{figure} 

\subsubsection{Reconstructing training and test data}
One argument in favor of character disentangled representations is that they should in principle be equally capable of representing any words comprised of the same characters, even if those words were not present during training. 
Recall that we trained on a set of valid three-letter English words. Using the same characters, we can construct test ``words'' that are not in the English dictionary and test the ability of our latent factor representations to reconstruct them. 
We visualize some words from the training and test set along with their reconstructions using MinSyn in Fig.~\ref{fig:word_recon}.
In Table~\ref{tab:result}, we compare the average $L_2$ loss on training and test data across methods. 
Minimizing synergy leads to the best character disentanglement according to ACC and, as hypothesized, the most disentangled method also produces the best reconstruction results on both training and test data.
Note that ICA and InfoMax use PCA to reduce to 9 components, before producing an orthogonal transformation into a basis of independent components, and therefore the reconstruction scores are identical even though the ACC's differ. 
We also found that MinSyn constrained representations were more robust to different types of out-of-domain noise than their unconstrained counterparts, but we omit those results here due to space.

\begin{table}[htbp]
  \label{tab:result}
  \centering
  \begin{tabular}{llll}
    \toprule
    Method     & Train loss     & Test loss & ACC\\
    \midrule
    MinSyn (Binary) 		&  \bf 12.94  & \bf 48.12  & \bf 0.558  \\
    MinSyn (Gaussian) 		&  58.68  &  91.26  &  0.670 \\
    ICA     				&  49.83  &  85.23 & 0.824 \\
    InfoMax     			&  49.83  &  85.23 & 0.849	\\
    PCA 					&  49.83  &  85.23 & 0.950 \\
    Auto-encoder 			& 35.98  & 73.67 & 1.07 \\
    Denoising 				& 38.26 & 76.40 & 1.08 \\
    \bottomrule
  \end{tabular}
  \vskip -0.2in
\end{table}

\section{Related work}\label{sec:related}

Decomposing information in terms of unique, synergistic, and redundant components is an active area of research ~\cite{james2017, williamsbeer, griffith}, with reviews of proposed measures appearing in~\cite{mpisynergy} and \cite{synergy_review}. 
While synergy is rarely mentioned with regard to representation learning, one motivation for exploring CI synergy was a new approach to latent factor models demonstrating state-of-the-art results for clustering and covariance estimation~\cite{blessing}; the approach combines total correlation explanation~\cite{nips2014,corex_theory}
with a condition that can be recognized in the context of this paper as a CI synergy constraint.

Similarly to the notion of synergy, disentanglement in representation learning has resisted a unique characterization, with statistical independence~\cite{ica,bell95} and sparsity via L1 regularization representing the most common approaches for achieving this goal \cite{lasso,bengioreview}.   
Another successful idea to reduce the fragility of learned representations involves adding noise to the latent factors (e.g. dropout). One can speculate that dropout reduces synergies in representations since it is not possible combine information from all latent factors (some fraction of which will be dropped out) to produce a given prediction.
A recent approach~\cite{informationdropout} generalizes dropout with learned multiplicative noise to approximate the Information Bottleneck objective \cite{tishby2000information}.  Disentangled \textit{and} minimally sufficient representations are encouraged by minimizing the objective $\sum_j I(X_{1:n}:Z_j) + TC(Z_{1:j})$. 
Finally, other efforts attempt to guide Generative Adversarial Networks toward learning disentangled representations by maximizing mutual information with latent codes that incorporate prior knowledge about the data~\cite{infogan}. 

A line of work which superficially contradicts this one is the ``Integrated Information Theory of Consciousness'' (IIT)~\cite{iit}. This line of work proposes that what makes consciousness (which we might call a learned neural representation) unique is precisely the degree to which the whole is more than the sum of the parts. A synergy measure is then introduced (integrated information), and consciousness is posited to maximize this synergy measure. While it appears that IIT proposes representations that maximize synergy while we propose representations that minimize synergy, these approaches actually address different ideas. IIT is concerned with the ``internal synergy'' of the brain as a system. In contrast, this work always focuses on the \emph{synergy with respect to a given prediction}. It is possible that a system might have high ``internal synergy'' as defined in IIT while having low synergy with respect to particular predictions, hence IIT and MinSyn are not directly comparable. 


\section{Conclusion}\label{sec:conclusion}


The absurdity of the Grue language mirrors a common dysfunction in machine learning. Many factors (grue/bleen and night/day) have to be combined to describe simple observations (visual appearance of an object) and therefore individual factors (like grue) do not seem very meaningful to us. Concepts like blue/green seem more natural and we suggest that informational synergy provides a way to quantify this intuition. 
This presents an opportunity to leverage the long history of academic interest in synergy measures to quantify and eliminate undesirable and unnecessary synergies in representation learning, hopefully leading to more interpretable models.

This work provides a first step in exploring the promise of reducing synergies in learning, however a number of challenges remain. 
Decades of research on synergy measures have not produced a consensus so finding the best measure for learning is an open problem. Because this field has remained theoretical, even popular synergy measures are difficult to estimate with little effort expended on tractable approximations. Existing work focuses on identifying synergies in observed data, while we suggest a fundamental shift in focus to using synergy as a principle for designing representations. Finally, while we minimized synergy in this work, it could be that reducing synergies with a tunable regularization strength would be more effective and that some amount of synergy is desirable or even necessary in some cases. 

Despite many open questions, we showed that at least one measure, CI synergy, is tractable and straightforward to apply to representation learning. We showed that MinSyn led to the learning of qualitatively different representations. 
We presented a benchmark task for disentangling factors of variation in data and and found that MinSyn learning succeeded where other approaches failed. 
While this proof of concept is encouraging, the difficulties we highlighted indicate ample room for improvement. 



\appendix

\subsection{Derivation of GK Synergy for the Gaussian Case}
\label{sec:appendix_gk}
\paragraph{GK synergy for Gaussians}  
We have seen that measuring GK synergy depends on a difficult optimization problem to find the union information.  However, if we restrict ourselves to Gaussian distributions, we are able to obtain an analytic expression for GK synergy. We begin by recasting the optimization in Eq.~\ref{eq:union} after restricting ourselves to Gaussian distributions. The scaled and centered Gaussian distributions are completely characterized by a covariance matrix, $M$, which must be positive semi-definite. We take the sub-matrix that is the covariance only for $Z_{1:n}$ to be $\Sigma$. The correlations between $Z_i$ and $X$ are $\rho_i$. The formula for mutual information is expressed in terms of the determinants of these matrices, with marginal constraints imposed on the covariance. 
\begin{equation}\label{eq:mi_opt}
\begin{aligned}
U(Z_{1:n};X) =~&  \underset{M \succeq 0}{\text{min}}
& & \half \log \frac{\det \Sigma}{\det M} \\ 
&  \text{s.t.}
& & M_{ii}=1, M_{i,n+1}=\rho_i, \forall i
\end{aligned}
\end{equation} 
Using the Schur complement, we can re-write the objective as $-\half \log(1 - \rho^\top \Sigma^{-1} \rho)$ which is equivalent to minimizing $\rho^\top \Sigma^{-1} \rho$. We rewrite the minimization involving a matrix inverse using a trick. We construct a matrix, $\bar M$, which is the same as $M$ but with $\bar M_{n+1,n+1}=t$.  
\begin{align}\label{eq:primal}
 \underset{\bar M \succeq 0}{\text{min}}~ t \quad \text{s.t.}\quad \bar M_{ii}=1, \bar M_{i,n+1}=\rho_i, i=1,\ldots,n
\end{align} 
Looking at the Schur complement for the condition $\bar M \succeq 0$, we see that minimizing $t$ is equivalent to minimizing $\rho^\top \Sigma^{-1} \rho$. 
Now that we have a canonical SDP we can use standard methods to construct a dual program. Feasible points of the dual provide a lower bound on the solution to our primal problem, Eq.~\ref{eq:primal}, and vice versa~\cite{boyd}. If we find a feasible point for the dual and primal whose objective value coincides, then this point is the optimum for both programs. The dual is as follows. 
\begin{align*}
 \underset{u, v \in \mathbb R^n}{\text{min}}
~ \sum_i 2 v_i \rho_i - u_i 
\quad  \text{s.t.} \quad
\begin{pmatrix}
u_1 &\ldots 0 & -v_1 \\
\vdots &  ~~u_n& -v_n \\
-v_1  & -v_n& 1
\end{pmatrix}
\succeq 0 
\end{align*} 
To prove the optimum of the primal problem, we give a feasible point for both the dual and primal problems and show their objective values coincide. For the dual, let $k = \arg \max_i \rho_i$. Set all $v_i = u_i=0$, except $v_k=\rho_k, u_k = \rho_k^2$. This satisfies the constraints and gives an objective of $\rho_k^2$. 
For the primal program, we propose the following feasible point. 
\begin{align}\label{eq:feasible}
 \bar M = \begin{pmatrix}
\mathbb I_n & \rho \\
\rho^\top & \rho_k^2
\end{pmatrix}
+ B
\end{align}
The symmetric matrix $B$ is defined so that $\forall i = 1,\ldots,k-1,k+1,\ldots, n , B_{i,k} = \rho_i/\rho_k$, or zero otherwise.  Inspection of the matrix reveals that it is positive semidefinite and the value of the objective is $\rho_k^2$. 

Finally, we can write the expressions for union information and GK synergy in terms of the correlation, $\rho_k$, with the largest magnitude. 
\begin{align}
U(Z_{1:n};X) &= I(Z_k;X) = \half \log(1 - \rho_k^2) \\
S(Z_{1:n};X) & = \half \log\frac{1 - \rho^\top \Sigma^{-1} \rho}{1 - \rho_k^2}
\end{align}

If we now consider optimizing this quantity over some variational distribution $q$ matching the pairwise marginals,
$$
q_{GK}(z_{1:n},x) =\arg \min_{q(z_{1:n},x)} S(Z_{1:m};X) \quad  \text{s.t.} \quad p(z_i, x) = q(z_i,x) 
$$
we see that $U(Z_{1:n};X)$ and $I(Z_i;X)$ are both constant for given marginals (since they depend only on $\rho$). Therefore, GK synergy reduces to minimizing mutual information subject to the marginal constraints. 
While we cannot solve this problem in general, we can again turn to the special case of Gaussian distributions for a solution. This optimization is exactly the one we had to solve to calculate the union information in Eq.~\ref{eq:mi_opt}. We found and proved the optimal point using SDPs. The solution can be expressed in terms of the covariance matrix over $X$, $\Sigma$, which can be read from Eq.~\ref{eq:feasible}. For simplicity and w.l.o.g., we re-order the $\rho_i$ so that $\rho_1$ has the largest magnitude. In that case, $q_{GK}(z_{1:n})$ is defined by its covariance matrix as follows. 
\begin{align}\label{eq:gk_sigma}
\Sigma = 
\begin{pmatrix}
1 & \rho_2 / \rho_1 & ... & \rho_n/\rho_1 \\
 \rho_2 / \rho_1 & 1 & 0  & 0 \\
  \vdots & 0 & \ddots & 0 \\
   \rho_n / \rho_1 & 0 & 0 & 1 \\
\end{pmatrix}
\end{align}
As a sanity check, we can use Eq.~\ref{eq:gk_sigma} to calculate that the value of the minimum of $S(Z_1, Z_2;X)$ that appears in the plot in Fig.~\ref{fig:example} should be $\Sigma_{1,2} = \rho_2 / \rho_1 = \half / \threefour = \twothree$. 
A matrix in this special form is called an arrowhead matrix. The inverse can be found using standard identities\footnote{If more than one $\rho_i$ have the same, maximal magnitude, the inverse does not exist and the space of solutions is degenerate.} 
and is a diagonal matrix plus a rank one matrix. This inverse is needed to complete our analysis since for Gaussians, $q_{GK}(x|z) \sim \mathcal N(\rho^\top \Sigma^{-1} z, 1 - \rho^\top \Sigma^{-1} \rho)$.
\begin{align}\label{eq:gk_gauss}
q_{GK}(X|Z_{1:n}=z) &\sim \mathcal N\left(\rho_1 z_1, 1 - \rho_1^2\right),
\end{align}
After all this work, we come to the trivial solution that the least synergistic relationship according to MI synergy measures is to take the best predictor and throw the others away! This surprising conclusion holds for Gaussians, but not in the more general case, where no analytic form for GK synergy is known. 
In the Gaussian case, if we have many variables to predict, $X_i$, then minimizing GK synergy implies that each should depend on only a single latent factor. This is equivalent to a Gaussian latent tree model. The implications for learning for minimizing GK synergy in the general case remain an open question. 


\subsection{Implementation Details}\label{sec:implementation}
All models were implemented using TensorFlow \cite{tensorflow} and optimized using  Adam~\cite{adam} with a learning rate $0.001$.
\subsubsection{MinSyn Gaussian}
We consider an auto-encoder with learned encoder weights and biases $w,b$, and a softplus activation function $z_j = \text{softplus}(\sum_i w_{j,i} x_i + b_i)$, and decoder weights and biases $u, d$ according to the linear decoder $\bar x_i = \sum_j u_{i,j} z_j + d_j$, trained to minimize the expected squared reconstruction error, $\mathbb E [|X-\bar X|^2_2]$. The only difference between our implementation and a standard one is that instead of learning the decoder weights as separate parameters, they are set according to Eq.~\ref{eq:ci_gauss}:  $u_{i,j} = \frac{1}{1+R_i} \frac{\rho_{ij}}{1-\rho_{ij}^2}$. 

\subsubsection{MinSyn Binary}
In the binary case, we re-scale pixel grayscale values into the range $[0,1]$ and interpret them as probabilities for a binary random variable to be on or off. The encoded state, $Z_{1:m}$ is constructed as a sigmoid of a linear function of the inputs. The decoder is specified similarly, except with weights specified as in Eq.~\ref{eq:decoder}. We use the cross entropy loss as is typical for binary random variables.  

\subsubsection{Auto-encoders} 
For comparison, we also consider a fully-connected auto-encoder where the decoder weights are trained rather than fixed according to the CI-minimizing distributions above.  We use a softplus non-linearity in the encoder and a sigmoid activation in the decoder, training to minimize cross entropy loss.  For the denoising auto-encoder, we add Gaussian noise with $\sigma=.25$ to each input feature.

\section*{ACKNOWLEDGMENT}
We thank Jimmy Foulds, Virgil Griffith, and David Kale for comments on this project. The authors acknowledge support from from DARPA grants W911NF-16-1-0575 and FA8750-17-C-0106. 

\bibliographystyle{IEEEtran}
\bibliography{sections/gversteeg_bibdesk}

\end{document}